\documentclass{article}

\usepackage[nonatbib, preprint]{neurips_2025}

\usepackage[utf8]{inputenc} 
\usepackage[T1]{fontenc}    
\usepackage{hyperref}       
\usepackage{url}            
\usepackage{booktabs}       
\usepackage{amsfonts}       
\usepackage{nicefrac}       
\usepackage{microtype}      
\usepackage{dsfont}

\usepackage{tcolorbox} 
\usepackage{amsmath}
\usepackage{caption}

\usepackage{listings}
\usepackage{xcolor}

\lstdefinestyle{custompython}{
    language=Python,
    basicstyle=\ttfamily\small,
    breaklines=true,
    breakatwhitespace=true,
    breakautoindent=false,
    showlines=true,
    keywordstyle=\color{blue},
    commentstyle=\color{gray},
    stringstyle=\color{red},
    numbers=none,
    frame=single,
    captionpos=b
}

\title{EvoGrad: Metaheuristics in a Differentiable Wonderland}

\author{%
  Beatrice F.R.~Citterio\\
  Department of Computing Sciences\\
  Bocconi University\\
  Milan, 20136, Italy \\
  \texttt{beatrice.citterio2@studbocconi.it} \\
  \And
  Andrea Tangherloni \thanks{Corresponding author} \\
  Department of Computing Sciences\\
  Bocconi Institute for Data Science and Analytics\\
  Bocconi University\\
  Milan, 20136, Italy \\
  \texttt{andrea.tangherloni@unibocconi.it} \\
}

\begin{document}

\maketitle

\begin{abstract}
    Differentiable programming has revolutionised optimisation by enabling efficient gradient-based training of complex models, such as Deep Neural  Networks (NNs) with billions and trillions of parameters.
    However, traditional Evolutionary Computation (EC) and Swarm Intelligence (SI) algorithms, widely successful in discrete or complex search spaces, typically do not leverage local gradient information, limiting their optimisation efficiency.
    In this paper, we introduce EvoGrad, a unified differentiable framework that integrates EC and SI with gradient-based optimisation through backpropagation.
    EvoGrad converts conventional evolutionary and swarm operators (e.g., selection, mutation, crossover, and particle updates) into differentiable operators, facilitating end-to-end gradient optimisation.
    Extensive experiments on benchmark optimisation functions and training of small NN regressors reveal that our differentiable versions of EC and SI metaheuristics consistently outperform traditional, gradient-agnostic algorithms in most scenarios.
    Our results show the substantial benefits of fully differentiable evolutionary and swarm optimisation, setting a new standard for hybrid optimisation frameworks.
\end{abstract}

\section{Introduction}

Differentiable programming (DP) has emerged as a transformative computational paradigm \cite{blondel2024}, enabling end-to-end optimisation of computational models by leveraging Automatic Differentiation (AD) \cite{baydin2018automatic}.
Neural networks (NNs) represent a specific yet highly successful instance of DP \cite{scardapane2024alice}, as they are composed of structured sequences of differentiable primitives that operate on tensors.
Each primitive is defined explicitly by input/output signatures, enabling compositionality and modularity.
The possibility of combining subsequent differentiable primitives, combined with efficient gradient computation via backpropagation, facilitates end-to-end training and scalability, underpinning the remarkable performance and great success of NNs in multiple domains like computer vision, natural language processing, and scientific modelling.

While DP excels in smoothly differentiable domains, many real-world optimisation problems (e.g., symbolic regression, neural architecture search, combinatorial design, and robotic morphology) involve discrete, hybrid, or structurally complex search spaces that are difficult or impossible to optimise with gradient-based methods alone.
Evolutionary Computation (EC) \cite{back1997handbook}, characterised by population-based global search strategies, and Swarm Intelligence (SI) \cite{eberhart2001swarm}, inspired by collective behaviours in nature, naturally complement DP by providing robust optimisation in non-differentiable or discontinuous search spaces.
EC methods, such as Genetic Programming (GP) \cite{koza1994genetic}, Genetic Algorithms (GA) \cite{mitchell1998introduction}, and Evolutionary Strategies (ES) \cite{back1997handbook}, and SI techniques, such as Particle Swarm Optimisation (PSO) \cite{kennedy1995particle}, excel at handling discrete structures, nonlinear interactions, and combinatorial spaces, navigating black-box landscapes without relying exclusively on gradient information.
However, classical EC and SI methods typically do not consider the rich local gradient information that is available within the differentiable parts of the search spaces, limiting their efficiency and scalability.
Recent works have begun bridging this gap through gradient-aware evolutionary algorithms, where differentiable relaxations and hybrid gradient components are introduced. 

Early work in differentiable GP (e.g., Differentiable Cartesian Genetic Programming (dCGP)) \cite{izzo2016} and exact symbolic derivatives for GP \cite{rockett2019d} paved the way for incorporating AD into symbolic regression.
The authors introduced automatic differentiation within GP trees, enabling efficient tuning of numerical constants and opening new possibilities like symbolic regression and differential equation solving.
Zeng \textit{et al.} \cite{zeng2023differentiable} and Anthes \textit{et al.} \cite{anthes2025transformer} extended these ideas using continuous symbolic tree relaxations and Transformer-based semantic operators, respectively, enabling gradient-informed evolution within high-dimensional symbolic domains.
In neural architecture search (NAS), hybrid approaches such as SpiderNet \cite{geada2022spidernet}, CGP-NAS \cite{garciagarcia2023continuous}, and dCGPANN \cite{martens2019neural} combine discrete evolutionary exploration with differentiable surrogates or relaxed encodings, enabling rapid discovery of efficient neural models.
Gradient-based methods have also been integrated into evolutionary strategies themselves: GENNES \cite{faury2019} uses generative neural networks to parameterise ES sampling distributions, while Evolution Transformer \cite{lange2024evolution} meta-learns update rules via Transformer architectures for in-context generalisation across tasks, showing remarkable generalisation capabilities across diverse black-box tasks.
Similarly, in PSO, hybrid algorithms combining global particle exploration with local gradient descent (GD) methods have shown improved convergence in continuous settings, e.g., for photovoltaics estimation \cite{zulu2023efficient} and neural control applications \cite{ejigu2022gradient}.
These strategies use PSO for exploration and GD for fine-tuning, demonstrating the synergy between swarm and gradient-based optimisation.
In evolutionary robotics and morphological optimisation, integrating differentiable physics simulation into EC strategies has yielded substantial improvements.
Differentiable physics has also catalysed innovation in evolutionary robotics.
Kurenkov and Maksudov reduced sample complexity in policy search via gradients from simulators \cite{kurenkov2021}, Horibe \textit{et al.} evolved soft robots with gradient-based damage recovery \cite{horibe2021regenerating}, and Strgar \textit{et al.} evolved entirely differentiable robotic morphologies for smoother training and fabrication \cite{strgar2024evolution}.

Despite these advancements, existing approaches typically incorporate gradients as auxiliary signals, with EC and SI-based processes predominantly remaining discrete, hand-crafted, and separated from gradient-based optimisation routines.
There remains a lack of a unified, fully differentiable EC and SI optimisation framework where all components (e.g., representation, selection, combination) are integrated within a differentiable computation graph.

In this paper, we present EvoGrad, a novel framework that fully integrates DP, EC, and SI.
EvoGrad reframes the entire optimisation process as a differentiable, parameterised computation:

\begin{itemize}
    \item \textbf{Differentiable Representation}: Individuals (particles, chromosomes) are learnable tensors parameterised within a differentiable model.
    \item \textbf{Differentiable Operators}: Mutation, crossover, and swarm operators are implemented as differentiable layers with learnable hyperparameters.
    \item \textbf{Differentiable Selection and Fitness Assignment}: Population selection is implemented via softmax and Gumbel–Softmax distributions, enabling backpropagation through parent selection and replacement.
\end{itemize}

To validate the robustness of EvoGrad, we implemented fully differentiable versions of CMAES, PSO, GA, and DE, using PyTorch and a shared optimisation loop.
We carefully reproduced canonical operator behaviour (e.g., crossover, mutation) and introduced pathwise gradients for sampling, Gumbel–Softmax relaxations for selection, and Binary-Concrete masks for stochastic operations.
All hyperparameters are fully learnable, and no restarts or hand-tuned annealing schedules are required.
To show the viability of EvoGrad, we performed experiments on standard continuous optimisation benchmarks (e.g., Ackley, Michalewicz, Rosenbrock, Griewank).
Our goal is not to achieve state-of-the-art performance, but to show that differentiable versions can match or exceed the performance of their classical counterparts, while offering the benefit of adaptive hyperparameter learning, smooth integration with neural components, and compatibility with gradient-based meta-learning pipelines.
It is important to note that all the algorithms in this study are designed for single-objective constrained optimisation.


\section{Methods}

We start this section by revisiting the canonical formulations of four widely used population-based optimisation algorithms: PSO \cite{kennedy1995particle}, GAs \cite{mitchell1998introduction}, Differential Evolution (DE) \cite{storn1997differential}, and the Covariance Matrix Adaptation Evolution Strategy (CMAES) \cite{hansen2001completely}.
These algorithms serve as the foundation for a wide range of global optimisation techniques due to their clarity, robustness, and ability to navigate complex, multimodal search landscapes.
Despite their success, classical implementations treat the optimisation process as a black box, preventing the use of gradient information and the integration with differentiable pipelines such as gradient-based meta-learning.
To address this limitation, we present differentiable counterparts for each algorithm by reformulating their stochastic operators using reparameterised (pathwise) transformations.
Our versions enable continuous and differentiable control over the algorithm's hyperparameters (e.g., inertia factor, mutation scale, crossover rate), allowing gradients of the loss function to flow through the optimiser itself.
In doing so, our algorithms exploit gradient-based adaptation to automatically set their hyperparameters throughout the entire optimisation process, bridging the gap between population-based and gradient-based optimisation.
Finally, we detail a common optimisation loop where the entire population-based search and meta-learning is viewed as a differentiable computational graph.
This generalised loop encompasses the four methods above under a shared interface, making it straightforward to switch between algorithms or extend them with task-specific components.

\subsection{Classical population-based optimisers}

We consider the general problem of global continuous optimisation: given an objective function \( f: \mathbb{R}^d \to \mathbb{R}\), we seek to identify a solution \(\mathbf{x}^\star \in \mathbb{R}^d \) such that \( \mathbf{x}^\star = \arg\min_{\mathbf{x} \in \mathcal{X}} f(\mathbf{x})\),
where \( \mathcal{X} \subseteq \mathbb{R}^d \) is a bounded, typically box-constrained search domain:
\(
\mathcal{X} = \left\{ \mathbf{x} \in \mathbb{R}^d \;\middle|\; \mathbf{l} \leq \mathbf{x} \leq \mathbf{u} \right\},
\)
with \(\mathbf{l}, \mathbf{u} \in \mathbb{R}^d\) denoting the coordinate-wise lower and upper bounds.

In settings where \( f \) is non-convex, multimodal, or non-differentiable, classical gradient-based methods often fail or require extensive tuning.
Population-based metaheuristics (e.g., PSO, GAs, DE, and CMAES) have emerged as robust alternatives that rely on stochastic sampling, selection, and variation operators rather than gradient information.
These methods evolve a population of candidate solutions across discrete generations, balancing exploration (diversity across the search space) with exploitation (local refinement of promising regions).

Let \(\mathbf{X}^{(t)} = \left\{ \mathbf{x}_i^{(t)} \in \mathbb{R}^d \right\}\), with \(i=1, \dots, N\), denote the population of size \(N\) at generation \(t\), and let \(f(\mathbf{x})\) be a deterministic or stochastic objective that assigns a scalar fitness to each individual \( i \).
We now briefly summarise the four classical algorithms we later differentiate.

\subsubsection{Particle Swarm Optimisation (PSO)}
PSO models each individual (i.e., particle) as a point \(\mathbf{x}_i \in \mathbb{R}^d\) moving through the search space under the influence of two main attractors: the cognitive (i.e., personal memory) and social (i.e., the best individual found so far) factors.
Each particle maintains a velocity vector \(\mathbf{v}_i \in \mathbb{R}^d \), updated at each iteration via:
\[
\begin{aligned}
\mathbf{v}_i^{(t+1)} &=
\omega \mathbf{v}_i^{(t)} + c_1 \mathbf{r}_1 \odot (\mathbf{p}_i - \mathbf{x}_i^{(t)}) + c_2 \mathbf{r}_2 \odot (\mathbf{g} - \mathbf{x}_i^{(t)}), \\
\mathbf{x}_i^{(t+1)} &= \mathbf{x}_i^{(t)} + \mathbf{v}_i^{(t+1)},
\end{aligned}
\tag{1}
\]
where \( \omega \) is the inertia coefficient, \(c_1, c_2 \) are the cognitive and social acceleration factor, respectively, \( \mathbf{r}_1, \mathbf{r}_2 \sim \mathcal{U}(0,1)^d \) are random numbers, \( \mathbf{p}_i \) is the personal best position of particle \( i\) found so far, \( \mathbf{g} \) is the global best found by the swarm so far, and \( \odot \) denotes the element-wise multiplication.

The velocity vectors \(\mathbf{v}_i\) are optionally clamped between \(\mathbf{v}_{min}\) and \(\mathbf{v}_{max}\), while the individuals \(\mathbf{x}_i\) are clamped or bounced at boundaries to enforce \(\mathbf{x}_i^{(t)} \in \mathcal{X}\).

\subsubsection{Genetic Algorithms (GAs)}

Real-valued GAs evolve a population of candidate solutions through stochastic parent selection (e.g., roulette or tournament), recombination (crossover), and mutation.
Given at least two parents, selected using any selection strategy, an offspring \( \mathbf{c} \in \mathbb{R}^d \) is generated component-wise by using a crossover strategy, such as blend crossover and SBX (Simulated Binary Crossover) \cite{deb1995simulated}.
In most GA implementations, the crossover operator is applied with a per-individual crossover rate \(c_r  \in[0,1]\).
Thus, with probability \(c_r\), crossover is performed (e.g., blend crossover or SBX).
With probability \(1 - c_r\), the offspring is a direct copy of one parent (often the better one).

\textbf{Blend crossover:}
Given two parents \(\mathbf{p}, \mathbf{q} \in \mathbb{R}^d\), the offspring is generated as:
\(
\mathbf{c} = \alpha \mathbf{p} + (1 - \alpha) \mathbf{q}, \; \alpha \sim \mathcal{U}(0,1)
\)

\textbf{SBX (Simulated Binary Crossover):}
Given two parents \(\mathbf{p}, \mathbf{q} \in \mathbb{R}^d\), the offspring is generated as:
\[
c_j = \frac{1}{2}[(1 + \beta_j)p_j + (1 - \beta_j)q_j], \quad
\beta_j =
\begin{cases}
(2u_j)^{1/(\eta_c + 1)}, & u_j < 0.5, \\
[1 / (2(1 - u_j))]^{1/(\eta_c + 1)}, & u_j \geq 0.5,
\end{cases}
\]
where \(u_j \sim \mathcal{U}(0,1)\), and \(\eta_c > 0\) controls the shape.
Note that some implementations also apply the crossover per gene (i.e., coordinate) rather than per individual.

After the crossover, a mutation operator is applied to the generated offspring to introduce stochastic perturbations that promote genetic diversity and enable local exploration around the recombined solution.

\textbf{Gaussian mutation:}
The Gaussian mutation perturbs a generated offspring \( \mathbb{c} \) as:
\(
\mathbf{c} = \mathbf{c} + \boldsymbol{\epsilon}, \quad \boldsymbol{\epsilon} \sim \mathcal{N}(\mathbf{0}, \sigma^2 \mathbf{I}).
\)
Note that Gaussian mutation is typically applied to each coordinate independently.

\textbf{Polynomial Mutation:}
The polynomial mutation perturbs each gene (i.e., coordinate) of the generated offspring \( \mathbb{c} \) with a probability (i.e., mutation rate) \(m_r\), applying a smooth, bounded change controlled by a distribution index \(\eta_m > 0\) \cite{deb2014analysing}.
Let \(c_j \in [l_j, u_j]\) be the current value, and \(\delta_j\) the mutation offset.
Then:
\(
c_j^\prime = c_j + \delta_j (u_j - l_j),
\)
where \( \delta_j \) is computed as:
\[
\delta_j =
\begin{cases}
(2u)^{1/(\eta_m+1)} - 1, &\text{if } u < 0.5, \\
1 - [2(1 - u)]^{1/(\eta_m+1)}, &\text{otherwise},
\end{cases}
\quad
u \sim \mathcal{U}(0,1).
\]
Note that the mutation is applied to the coordinate \( j \) with probability \(m_r\); otherwise, \(c_j\prime = c_j\).

Finally, the elitism strategy can be used to ensure the survival of the best individual through the generations.

\subsubsection{Differential Evolution (DE)}
DE evolves each individual using mutation and binomial crossover.
The most used mutation strategies are \emph{DE/rand/1} and \emph{DE/current-to-best/1}.

\textbf{DE/rand/1:}
For each individual \(\mathbf{x}_{i}\), three randomly chosen distinct individuals 
\(\mathbf{x}_{r_1}, \mathbf{x}_{r_2}, \mathbf{x}_{r_3}\) are selected to create a donor vector \(\mathbf{v}_i\):
\(
\mathbf{v}_i = \mathbf{x}_{r_1} + F \cdot (\mathbf{x}_{r_2} - \mathbf{x}_{r_3}),
\)

\textbf{DE/current-to-best/1:}
For each individual \(\mathbf{x}_{i}\), two randomly chosen distinct individuals \(\mathbf{x}_{r_1}, \mathbf{x}_{r_2}\) are used to compute the donor vector \(\mathbf{v}_i\):
\(
\mathbf{v}_i = \mathbf{x}_i + F \cdot (\mathbf{x}_\text{best} - \mathbf{x}_i) + F \cdot (\mathbf{x}{r_1} - \mathbf{x}{r_2}),
\)
where \( \mathbf{x}_\text{best} \) is the current best individual.

\textbf{Binomial crossover:}
After that, the binomial crossover is applied to combine the individual \(\mathbf{x}_{i}\) and the donor vector \(\mathbf{v}_i\).
\[
u_{ij} =
\begin{cases}
v_{ij}, & \text{if } \tau_{ij} < \text{CR} \text{ or } j = j_{\text{rand}}, \\
x_{ij}^{(t)}, & \text{otherwise},
\end{cases}
\qquad \tau_{ij} \sim \mathcal{U}(0,1),
\]

where \(F \in (0, 2]\) is the mutation scale factor and \(CR\) is the crossover rate.
The trial vector  \( \mathbf{u}_i \) competes against the current parent \( \mathbf{x}_i^{(t)} \), and the fitter individual survives.

Note that classic DE does not implement \textit{elitism} in the traditional sense, but it implicitly preserves better individuals via its greedy, pairwise selection step.
This strategy guarantees that the population does not worsen locally (i.e., no individual becomes worse), locally preserving the fitness.
In EC, elitism typically refers to globally retaining the best-so-far solution, regardless of which individual it replaces. For example, in a \((\mu+\lambda)\) scheme, the best \(\mu \) individuals are selected from the union of parents and offspring.
\( \mu \) and \( \lambda \) are classical notations that refer to the number of parents and offspring, respectively.
In explicit elitism, the best global solution \(\mathbf{x}^\star \) is copied into the next generation even if it would otherwise be lost.
DE does not guarantee that the best solution across the whole population survives, unless it happens to win its local comparison.

\subsubsection{Covariance Matrix Adaptation Evolution Strategy (CMAES)}
CMAES models the population as samples from a multivariate Gaussian distribution \( \mathcal{N}(\boldsymbol{\mu}, \sigma^2 \mathbf{C}) \), with dynamic adaptation of the mean, step-size, and full covariance matrix \( \mathbf{C} \in \mathbb{R}^{d \times d} \).
In each generation, the following sampling strategy is applied:
\(
\mathbf{x}_i^{(t)} = \boldsymbol{\mu}^{(t)} + \sigma^{(t)} \mathbf{L}^{(t)} \mathbf{z}_i, \qquad \mathbf{z}_i \sim \mathcal{N}(\mathbf{0}, \mathbf{I}),
\)
where \( \mathbf{L} \) is the Cholesky decomposition of \( \mathbf{C} \).
After evaluating all individuals, the mean \( \boldsymbol{\mu} \) is updated using a weighted recombination:
\(
\boldsymbol{\mu}^{(t+1)} = \sum_{i=1}^{\mu} w_i \mathbf{x}_{i:\lambda}^{(t)},
\)
with \( \mu < \lambda \) and weights \( w_i > 0 \).
\(\boldsymbol{w}\) is the vector of positive recombination weights summing to 1.
The evolution paths \(\mathbf{p}_\sigma \) and \( \mathbf{p}_c \) track cumulative steps for step-size control and rank-one updates, respectively.
The covariance \( \mathbf{C} \) is adapted as:
\(
\mathbf{C}^{(t+1)} = (1 - c_1 - c_\mu)\mathbf{C}^{(t)} + c_1 \mathbf{p}_c \mathbf{p}_c^\top + c_\mu \sum_{i=1}^\mu w_i \mathbf{y}_i \mathbf{y}_i^\top,
\)
with \(\mathbf{y}_i = \mathbf{L}^{-1}(\mathbf{x}_i - \boldsymbol{\mu})/\sigma \), and step-size \( \sigma \) updated via cumulative step-size adaptation (CSA):
\(
\sigma^{(t+1)} = \sigma^{(t)} \cdot \exp\left(\frac{c_\sigma}{d_\sigma} \left( \frac{\|\mathbf{p}\sigma\|}{E\|\mathcal{N}(\mathbf{0},\mathbf{I})\|} - 1 \right)\right).
\)

\subsection{The proposed differentiable algorithms}
The key step in turning a population-based optimiser into a corresponding differentiable version is to rewrite every random draw as a deterministic transformation of an external noise variable that is independent of the algorithm's parameters.
To do so, we can use a \textbf{pathwise} or \textbf{reparameterisation trick} that converts a random draw that depends on the model parameters into a deterministic, differentiable transformation of an auxiliary noise source that does not depend on those parameters.
Let \(\mathbf{x}\sim p_\theta(\mathbf{x}) \) be a random variable whose
distribution depends on model parameters \( \theta \).
The pathwise trick rewrites the sample as a deterministic transformation of parameter-free noise \( \boldsymbol\varepsilon: \mathbf{x}=g_\theta(\boldsymbol\varepsilon), \;
\boldsymbol\varepsilon\sim p(\boldsymbol\varepsilon)\) that is independent of \(\theta\).
With this re-parameterisation the Monte-Carlo estimate \( L(\theta)=f(\mathbf{x}) \) becomes \( \tilde L(\theta)=f\!\bigl(g_\theta(\boldsymbol\varepsilon)\bigr)\), which is
differentiable with respect to \( \theta \).
Automatic differentiation therefore produces an \textit{unbiased gradient estimate} \( \nabla_\theta L \approx \nabla_\theta \tilde L \).

Many operators (e.g., mutation masks, DE crossover) involve Bernoulli random variables \(m\in\{0,1\}\) that are not differentiable.
We replaced each hard bit \(m\in\{0,1\}\) by a relaxed random variable
drawn from the \textbf{Gumbel–Sigmoid} distribution 
\( \tilde m
= \sigma\!\left(\frac{1}{\tau} \bigl(\log u - \log(1-u) + \alpha\bigr)\right), \; u\sim\mathcal U(0,1), \alpha\in\mathbb R\), where \(\tau>0\) is the temperature, \( \alpha \) is a learnable logit, and \(\sigma(\cdot) \) the logistic function \cite{geng2020does}. 
To retain the discrete semantics at run-time, we apply the straight-through (ST) estimator.
During the forward pass, we apply a hard threshold \(m=\mathbb{I}[\tilde m>0.5]\) to preserve the exact discrete behaviour of the algorithm (e.g., decide whether a gene mutates).
During the backward pass, we use the straight-through estimator, pretending that the operation was continuous, \(\frac{\partial L}{\partial\alpha}= \frac{\partial L}{\partial\tilde m} \cdot \frac{\partial\tilde m}{\partial\alpha}\)., allowing gradients to update the learnable logit parameter \(\alpha\). 
Thus, uniform variates can now be expressed using a Binary-Concrete or Gumbel–Sigmoid transformation \cite{geng2020does}.
In contrast, Gaussian variates are obtained by multiplying a standard normal tensor by a learnable scale or covariance factors \cite{kingma2015}.

Parent selection in DE and tournament selection in GAs can be viewed as sampling from a categorical distribution rather than performing independent Bernoulli draws.
We therefore employ the Gumbel–Softmax relaxation \cite{jang2017categorical}:
\(
    \tilde{\mathbf p} =
    \operatorname{softmax}\!\Bigl(\frac{1}{\tau}\bigl(\log\mathbf u - \log(1-\mathbf u) + \boldsymbol\alpha\bigr)\Bigr),
    \;
    \mathbf u\sim\mathcal U(0,1)^N,
    \;
    \boldsymbol\alpha\in\mathbb R^{N},
\)
where \(\tau\) is the temperature and \(\boldsymbol\alpha\) a vector of learnable logits.
During the forward pass, it is possible to take the \texttt{argmax} of \(\tilde{\mathbf p}\) (straight-through) to obtain a hard parent index or use the soft selection (in our tests, we used the soft selection); during the backwards pass, gradients flow through the softmax, enabling the optimiser to shape the probability of selecting each individual.
This mechanism complements the Gumbel–Sigmoid masks described and provides a fully differentiable substitute for discrete sampling in all parent-selection steps.
In this way, the loss \(L=f(\mathbf x^{(t)})\) becomes a smooth function of both the search parameters (e.g., particle positions, population mean, step–size) and the \textit{meta-parameters} or \textit{hyperparameters} (e.g., inertia, scale factor, crossover rate, covariance, mutation index).
Autodiff can therefore propagate \( \partial L / \partial\theta \) through the entire sampling path.

\subsubsection{The unified gradient and exploration loop}

\begin{figure}[t]
    \centering
    \begin{lstlisting}[style=custompython, caption={\textsc{EvoGrad}: unified differentiable optimisation loop}, label={listing:optimisation_loop}]
    while algorithm.n_evals < max_evals:
        
        algorithm.optimiser.zero_grad(set_to_none=True)
    
        # Meta-heuristic steps and best fitness (scalar)
        loss = algorithm()
    
        # Gradients through sampling and meta-params
        loss.backward() 
    
        # Optimiser on all the learnable parameters (exploitation / hyper-learning step)
        algorithm.optimiser.step()                         
        # Commit the evolutionary changes
        algorithm.update_state()
    \end{lstlisting}
\end{figure}


In each generation, the following three phases are applied (see the code reported in \ref{listing:optimisation_loop}):
\begin{enumerate}
    \item Sampling: the algorithm draws a population through the differentiable operators described above and evaluates the objective, returning the best fitness as a scalar loss;
    \item Exploitation / hyper-learning: \texttt{loss.backward()} populates gradients with respect to all learnable quantities; a single gradient-descent based step adapts the individuals and meta-parameters in the direction that most reduces the loss.
    \item Commit: the state update (i.e., \texttt{algorithm.update\_state()}) copies the candidate tensors (still connected to the graph) into the persistent state.
    By interleaving exploration (stochastic sampling) with exploitation (outer-loop gradient step), EvoGrad combines global search with local, gradient-driven adaptation, while remaining entirely differentiable and compatible with standard autodiff.
\end{enumerate}
In all four algorithms, we store the current population itself as learnable parameters (i.e., PyTorch \texttt{nn.Parameters}).
Concretely, the matrix \(\mathbf{X}^{(t)}=\bigl[\mathbf{x}_1^{(t)},\ldots,\mathbf{x}_N^{(t)}\bigr]^{\!\top}\in\mathbb R^{N\times d}\) is randomly initialised once and registered as a parameter; hence it receives gradients during the backpropagation and is updated by the outer gradient-based optimisation step alongside the meta-parameters (e.g., \(\omega, \sigma, F,\eta_c\)).
This design removes the conceptual separation between ``solution space'' and ``parameter space''; indeed, the optimiser learns where to move each individual and how to adjust its own search operators within the same autodiff graph.

\subsubsection{Differentiable PSO}
For PSO, we retain the classical velocity update but treat the inertia factor \(\omega \) and the cognitive and social coefficients (\(c_1, c_2\)) as \texttt{nn.Parameters}.
We went one step further and assigned an individual inertia weight and social cognitive coefficients to every particle.
Formally, let \(\boldsymbol\omega, \mathbf c_1, \mathbf c_2\) denote \(N\)-dimensional vectors in \(\mathbb{R}^N\) of per-particle coefficients. 
All three vectors \(\omega, \mathbf c_1, \mathbf c_2\) are themselves \texttt{nn.Parameters}, allowing different particles to learn distinct exploration–exploitation profiles purely from gradient feedback.
Thus, the velocity update therefore becomes fully differentiable and is equal to \(\mathbf v_i^{(t+1)}=\omega_i\,\mathbf v_i^{(t)}
+ c_{1,i}\,\mathbf r_1\odot(\mathbf p_i-\mathbf x_i^{(t)})
+ c_{2,i}\,\mathbf r_2\odot(\mathbf g  -\mathbf x_i^{(t)})\),
with \(\mathbf r_1, \mathbf r_2 \sim\mathcal U(0,1)^d\) generated with the pathwise trick.

\subsubsection{Differentiable GAs}
Each individual \(\mathbf{x}_i\in\mathbb R^{d}\) is a learnable \texttt{nn.Parameter}; the population therefore receives gradients exactly as in PSO.
Regarding the crossover operators, we propose a differentiable version of SBX.


\textbf{SBX:} component-wise spreads are generated by the pathwise form \(\beta_j = (2u_j)^{1/(\eta_c+1)}\) if \(u_j<0.5\) else \(\beta_j =[2(1-u_j)]^{-1/(\eta_c+1)} \) with \(u_j\sim\mathcal U(0,1)\), where the index \(\eta_c\) is stored as \(\log\eta_c\) and learned.

Mutation is applied with a Binary-Concrete mask per gene.
If a gene is active, it can be modified using a differentiable version of the polynomial mutation.

\textbf{Polynomial Mutation:} it adds \(\delta_j(u_j,\eta_m)\) where \(\eta_m\) and the per-gene probability logits are learned.

As both masks and noise are reparameterised, GA's crossover rate, mutation rate, and indices are all tuned by exploiting the backpropagation and GD-based optimiser.

\subsubsection{Differentiable DE}
Each parent vector is a learnable \texttt{nn.Parameter}, and all mutation coefficients are learnable.
We rewrote the \emph{DE/rand/1} and \emph{DE/current-to-best/1} mutation strategies to be differentiable.
In both cases, the scale factor \(F=\exp(\phi)\) is automatically adjusted by learning \(\log\phi\ \), stored as an \texttt{nn.Parameter}.

\textbf{DE/rand/1:} \( \mathbf v_i = \mathbf x_{r_1}+F(\mathbf x_{r_2}-\mathbf x_{r_3})\)

\textbf{current-to-best/1:}\(\mathbf v_i = \mathbf x_i +F(\mathbf x_{\mathrm{best}}-\mathbf x_i)+F(\mathbf x_{r_1}-\mathbf x_{r_2})\)

The parents are sampled with a Gumbel–Softmax, which can be seen as a differentiable selection routine.
Similar to what we have done for the GA's polynomial mutation, we used a Binary-Concrete mask with a learnable crossover-rate logit for the binomial crossover.
Greedy one-to-one replacement is retained; when the trial loses, we apply a straight-through estimator so the outer optimiser still receives a gradient signal.
To enforce elitism in DE, we keep track of the global best \(\mathbf{x}^\star\) and, after one-to-one replacement, \(\mathbf{x}^\star\) is injected into the new population by replacing the worst individual.

\subsubsection{Differentiable CMAES}
We followed the implementation proposed by Auger and Hansen with full evolution path updates \cite{auger2012tutorial}.
The mean \(\boldsymbol\mu\!\in\!\mathbb R^{d}\), log step-size \(\log\sigma\), and the lower-triangular Cholesky factor \(\mathbf L\) are all learnable \texttt{nn.Parameters}.
Sampling, thus, follows the exact reparameterised form
\(\mathbf x = \boldsymbol\mu + \sigma\mathbf L\mathbf z,\; \mathbf z\sim\mathcal N(0,\mathbf I)\).
Ranking is replaced by a softmax-with-temperature so that the recombination weights \(w_i\) become differentiable functions of the raw fitness values.
The indicator \(h_{\sigma}\) in the evolution-path update is smoothed with a logistic window.
The diagonal jitter \(\varepsilon\) added before each Cholesky factorisation is computed from the smallest non-zero entry of \(\mathbf L\) but detached from the graph, guaranteeing numerical stability without spoiling gradients.

\section{Experiments}

\begin{figure*}[t]
    \centering
    \includegraphics[width=0.9\textwidth]{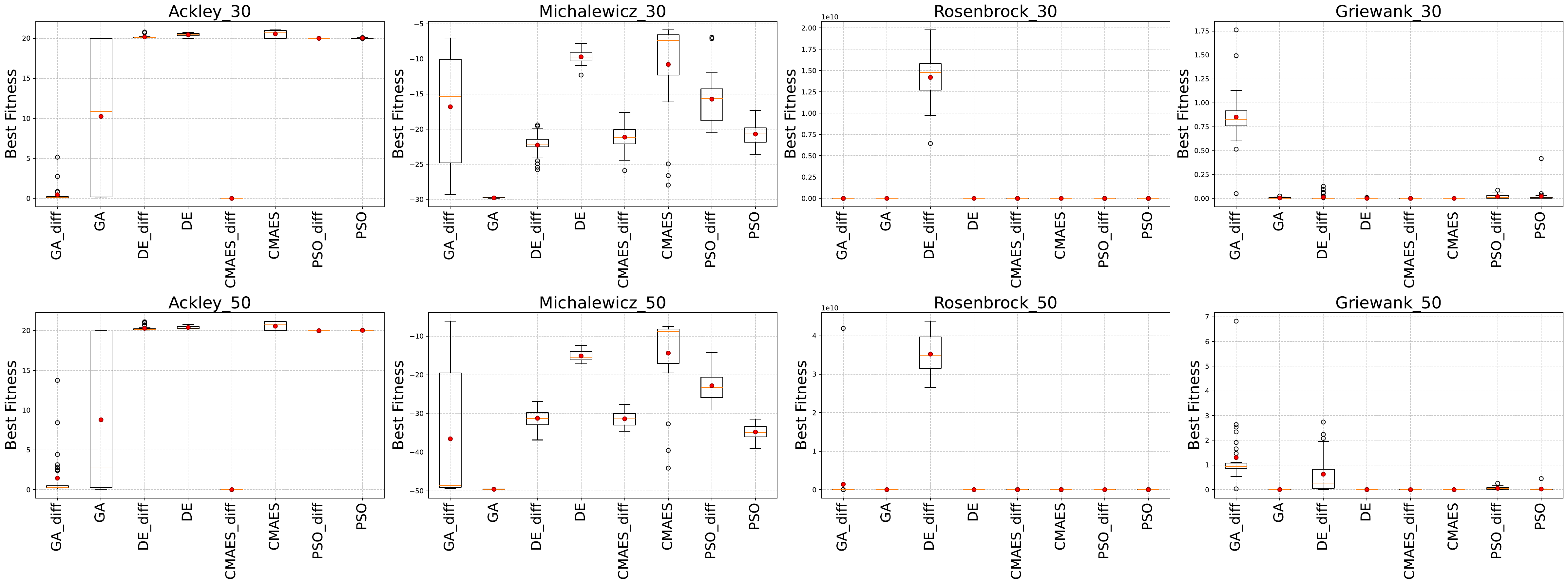}
    \caption{Final best fitness values achieved by each algorithm across standard continuous optimisation benchmarks: Ackley, Michalewicz, Rosenbrock, and Griewank.
    The top row reports results for 30-dimensional problems; the bottom row for 50-dimensional ones. 
    For each method (GA, DE, PSO, CMAES), we compare the classical version (plain) with our differentiable counterpart (diff).
    Each boxplot summarises the distribution over 30 independent runs, including median, interquartile range, and outliers.}
    \label{fig:benchmark_comparison}
\end{figure*}

To evaluate the performance of our differentiable metaheuristics, we performed experiments on four widely used continuous optimisation benchmarks: Ackley, Michalewicz, Rosenbrock, and Griewank.
These benchmarks span various challenges, including multimodality, narrow valleys, deceptive local minima, and non-convex curvature, and are standard in evolutionary optimisation.
We stress out that our primary goal is not to outperform state-of-the-art methods or tackle complex constrained problems like those proposed in well-known competitions (e.g., CEC benchmark suites), but to assess whether our differentiable reparameterisations of classic algorithms can match or obtain better performance than their non-differentiable counterparts under comparable settings.
We implemented only baseline versions of GA, DE, CMAES, and PSO, with no restarts or more advanced external heuristics.
For a fair comparison, we used the standard versions of the algorithms available in the \texttt{pymoo} library, selecting configurations that closely match our own implementations (e.g., disabling restarts and tuning the parameters to reflect canonical defaults).
For PSO, we employed the parameter-free variant included in \texttt{pymoo}, which requires no manual hyperparameter tuning.

For each benchmark function, we defined box constraints in \([-100, 100]^D\) and set the population size to 100 individuals, regardless of dimensionality.
Each algorithm was run for \(5000 \cdot D\) fitness evaluations.
We tested two dimensions: \(D = 30\) and \(D = 50\). 
Our differentiable variants were trained using the Adam optimiser with an initial learning rate of 0.01 and a \textbf{ReduceLROnPlateau} scheduler triggered every 100 generations (plateau factor = 0.5).
To perform all these tests, we used a workstation equipped with two Intel Xeon Gold 6152 CPUs and an NVIDIA RTX 4090 GPU.
All differentiable algorithms were implemented in PyTorch and executed on the GPU.
Notably, EvoGrad implementations run approximately \(3 \times\) faster than the equivalent \texttt{pymoo} versions executed on CPU.

Figure \ref{fig:benchmark_comparison} summarises the final best fitness achieved by each method over 30 independent runs, for both 30D and 50D settings. Boxplots show the distribution of final scores, providing insight into convergence behaviour and robustness of our approaches.

Our differentiable CMAES consistently achieves the best performance across nearly all benchmark functions and dimensions, particularly on Ackley, Griewank, and Rosenbrock, where it achieves both lower medians and tighter distributions.
This supports the effectiveness of adaptive covariance learning and step-size control in high-dimensional search spaces.
Differentiable PSO is also highly competitive, closely following CMAES in most settings.
This indicates that learning inertia and social/cognitive factors is effective.
Even though the \texttt{pymoo} PSO version is able to autotune its parameters, it often suffers from premature convergence, while its differentiable counterpart maintains a better exploration-exploitation balance.
In contrast, differentiable GA and DE perform comparably to their standard versions and occasionally underperform, particularly on Michalewicz and Rosenbrock, where gradient information may not align well with the problem geometry.
This suggests that gradient-aware recombination and mutation (e.g., SBX or polynomial mutation) benefit from smoothness but can mislead the search on rugged landscapes.
Rosenbrock, known for its narrow curved valley, remains challenging for all algorithms.
However, differentiable variants (particularly CMAES and PSO) manage to lower the loss more consistently, likely due to internal parameter learning.

\begin{figure*}[t]
    \centering
    \includegraphics[width=0.9\textwidth]{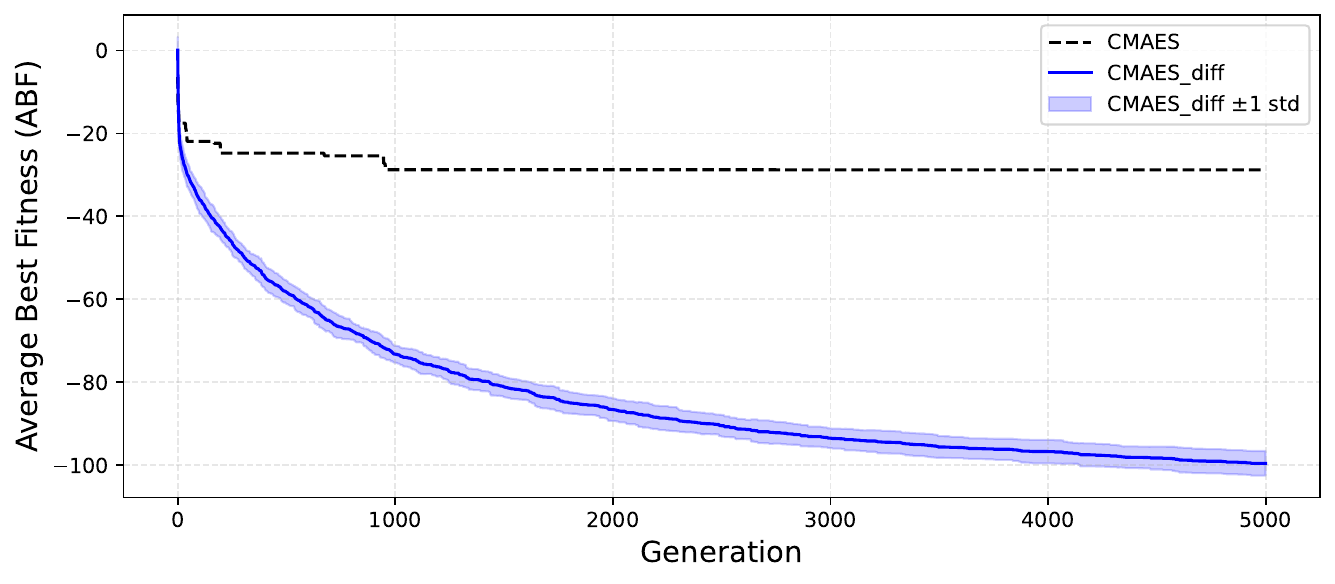}
    \caption{
    Convergence comparison between classical CMAES and our differentiable CMAES on the 500-dimensional Michalewicz function with a total budget of 500,000 fitness evaluations.
    We report the average best fitness (ABF) per generation. The CMAES\_diff curve represents the mean over 30 independent runs, with the shaded region indicating \(\pm 1\) standard deviation.
    The differentiable version consistently outperforms the classical baseline, demonstrating faster convergence and better final optima.
    }
    \label{fig:convergence}
\end{figure*}

To further evaluate the scalability and efficiency of our differentiable metaheuristics, we compared the classical and differentiable versions of CMAES on a high-dimensional instance of the Michalewicz function with 500 dimensions and a budget of 500,000 fitness evaluations.
While the standard CMAES implementation from the pymoo library required approximately 1 hour to complete a single run, our GPU-accelerated differentiable CMAES completed 30 independent runs in just 16 minutes on an NVIDIA 4090.
In terms of solution quality, the classical CMAES converged to a best fitness of -28.82, whereas the differentiable variant consistently achieved substantially lower values across all runs, with best fitness values ranging from -93.40 to -103.97 (see Figure \ref{fig:convergence}).
These results highlight the strong scalability and accelerated convergence of the differentiable formulation, particularly when optimising complex multimodal functions in high-dimensional search spaces.

To show the practical capabilities of EvoGrad beyond synthetic benchmarks, we evaluated its performance on a real-world regression task involving the Wine Quality (red) dataset \cite{cortez2009modeling}.
Specifically, we compared our differentiable CMA-ES against Adam (with default PyTorch settings) for training a small neural network.
The dataset comprises 11 input features and a perturbed version of the wine quality score as the regression target, where we added heteroscedastic noise using a log-normal distribution \(y = y + \mathrm{LogNormal}(0, 1)\). 
The NN architecture consisted of a single hidden layer with 128 neurons and tanh activation, followed by a linear output layer, for a total of 1,665 parameters.
All parameters were initialised uniformly in \([-10, 10]\).
To train the NN, we used the Mean Squared Error (MSE) as a loss function.
We trained the NN using 3000 epochs for Adam and 3000 fitness evaluations for CMAES (using a population size of 30).
Across 30 independent runs, Adam achieved an average best final loss (ABF) of \(71.90 \pm 15.34\) (min: 34.51, max: 89.80), while CMAES achieved a significantly lower ABF of \(5.66 \pm 0.39\) (min: 5.19, max: 6.47), with comparable computational time.
This highlights that EvoGrad, even without task-specific tuning, can outperform first-order gradient-based optimisers on noisy and discontinuous loss landscapes, further supporting its generality and robustness in challenging optimisation scenarios.

\section{Conclusions}
We introduced EvoGrad, a unified differentiable framework for classic population-based metaheuristics, including CMAES, PSO, GAs, and DE.
By re-parameterising each stochastic operator (e.g., mutation, crossover) and sampling strategy, we enabled end-to-end gradient-based learning over all algorithmic components, from the population to the algorithm's hyperparameters.
This differentiable reformulation turns traditionally black-box optimisers into fully differentiable modules, opening new opportunities for hybrid gradient-based and gradient-free optimisation.

Through a consistent pathwise gradient formalism, we showed that (\textit{i}) sampling from multivariate Gaussians (as in CMAES) can be expressed usingreparameterisedd Cholesky transforms, (\textit{ii}) crossover and mutation operators (e.g., SBX and polynomial mutation) can be relaxed with Binary-Concrete and Gumbel–Sigmoid variables, enabling differentiable masks, (\textit{iii}) parent selection, typically implemented via discrete tournaments or stochastic rankings, can be approximated using the Gumbel–Softmax trick, allowing soft, differentiable mixtures of individual, (\textit{iv}) inertia and social-cognitive parameters in PSO, and crossover/mutation rates in GA and DE, can be made learnable and optimised via backpropagation.
Our results show that differentiable metaheuristics can match or outperform their classical forms on standard continuous optimisation tasks.
Gradient-based learning of internal parameters like step-size, crossover rate, or velocity coefficients contributes to more efficient search behaviour. 
With EvoGrad, we reframed classical metaheuristics as differentiable modules, paving the way for meta-learning, adaptive search, and hybrid optimisation pipelines.

The main limitation of EvoGrad lies in the memory cost of unrolled optimisation, which is required to backpropagate through time across multiple generations.
While our meta-loss formulation circumvents this in the single-step case, full credit assignment over long horizons (e.g., in curriculum learning or reinforcement learning) will benefit from truncated backpropagation or learned meta-optimisers.
Another open question is how best to balance exploration and exploitation in this differentiable setting.
While classic algorithms rely on fixed heuristics (e.g., inertia decay or covariance adaptation), our framework allows learning these schedules, which may require new regularisers or control mechanisms to ensure stability and generalisation.
As future work, we are planning to extend EvoGrad to incorporate more advanced variants of DE and CMAES, like 
SHADE \cite{tanabe2013success} and BIPOP-CMAES \cite{hansen2009benchmarking}, as well as other complex metaheuristics such as Evolutionary Algorithm for COmplex-process oPtimization (EACOP) \cite{tangherloni2024modified}, with the goal of scaling differentiable metaheuristics to high-dimensional and noisy real-world optimisation problems.

\newpage
{
\small
\bibliographystyle{plain}
\bibliography{biblio}
}

\end{document}